\documentclass[11pt,a4paper]{article}
\usepackage[hyperref]{acl2020}
\usepackage{times}
\usepackage{latexsym}
\usepackage{hyperref}
\usepackage{url}
\usepackage{caption}
\usepackage{graphicx}
\usepackage{tabularx}
\usepackage{float}
\usepackage{booktabs}
\usepackage{multicol, multirow}
\usepackage{wrapfig}
\usepackage{subfigure}
\usepackage{enumitem}
\usepackage{color,soul}
\usepackage{longtable}
\usepackage{placeins}

\setul{0.6ex}{0.2ex}

\usepackage{microtype}

\newcommand{\DatasetName}{\textsc{CreativeSumm}}

\title{\DatasetName: Shared Task on \\ Automatic Summarization for Creative Writing}

\newcommand*{\affaddr}[1]{#1}
\newcommand*{\affmark}[1][*]{\textsuperscript{#1}}

\author{
  \textbf{Divyansh Agarwal}\affmark[$1$]
  \quad \textbf{Alexander R. Fabbri}\affmark[$1$] 
  \quad \textbf{Simeng Han}\affmark[$3$] 
  \quad \textbf{Wojciech Kry\'sci\'nski}\affmark[$1$] \\
  \quad \textbf{Faisal Ladhak}\affmark[$2$] 
  \quad \textbf{Bryan Li}\affmark[$5$] 
  \quad \textbf{Kathleen McKeown}\affmark[$2$] \\
  \quad \textbf{Dragomir Radev}\affmark[$3$] 
  \quad \textbf{Tianyi Zhang}\affmark[$4$] 
   \quad \textbf{Sam Wiseman}\affmark[$6$] \\
  \affaddr{\affmark[$1$]Salesforce Research} 
  \quad 
  \affaddr{\affmark[$2$]Columbia University}
  \quad 
  \affaddr{\affmark[$3$]Yale University}\\
  \quad 
  \affaddr{\affmark[$4$]Stanford University}
  \quad 
  \affaddr{\affmark[$5$]University of Pennsylvania} 
    \quad 
  \affaddr{\affmark[$6$]Duke University} \\
}

\date{}

\begin{document}
\maketitle
\begin{abstract}

This paper introduces the shared task of summarizing documents in several creative domains, namely literary texts, movie scripts, and television scripts. Summarizing these creative documents requires making complex literary interpretations, as well as understanding non-trivial temporal dependencies in texts containing varied styles of plot development and narrative structure. This poses unique challenges and is yet underexplored for text summarization systems. In this shared task, we introduce four sub-tasks and their corresponding datasets, focusing on summarizing books, movie scripts, primetime television scripts, and daytime soap opera scripts. We detail the process of curating these datasets for the task, as well as the metrics used for the evaluation of the submissions. As part of the \DatasetName~workshop at COLING 2022, the shared task attracted 18 submissions in total. We discuss the submissions and the baselines for each sub-task in this paper, along with directions for facilitating future work in the field.

\end{abstract}

\section{Introduction}\label{sec:introduction}

Contemporary research in text summarization systems has focused mainly on a select few domains such as news, scientific and legal articles. While summarizing these domains is important, the corresponding documents are limited in length and the diversity of the discourse structure.

There is a large body of creative texts available on the web that pose greater challenges for text summarization in NLP. These creative documents like books, movie scripts, and television scripts are usually written by `subject matter experts', are substantial in length and contain complex inter-dependencies between characters in the plotline~\cite{kryscinski2021booksum, Ladhak:20, Mihalcea:07}. Summarization of such text requires understanding many different genres, and offers the possibility of improving creating writing platforms~\cite{aparicio2016summarization, toubia2021poisson, gorinski2015movie}.

Such creative documents are often accompanied by a synopses, short-description or a summary, which allows readers to gauge their interest in the corresponding artifact.
Summaries of creative texts are widely used by students as study guides, and could be useful to experts that need to grade the quality of these documents for a particular audience. Furthermore, researchers may be interested in using the challenges posed by these creative texts, to identify the limitations of large language models at understanding complex discourse structure.  

To further summarization research of creative writing, we identified four particular domains, each with their own set of challenges. Building off of datasets recently released in the community, we develop a shared task, composed of four sub-tasks, and encouraged participants to further research in this direction.

For sub-task 1, we focus on summarization of chapters from literary texts like books, novels and stories. We curate a combined version of Booksum~\citep{kryscinski2021booksum} and NovelChapters~\citep{Ladhak:20} for this purpose. 

For sub-task 2, we use the Scriptbase dataset~\citep{gorinski2015movie}, which pairs movie transcripts with their corresponding Wikipedia summaries.

Sub-tasks 3 and 4 both relate to summarization of transcripts for TV shows, both derived from SummScreen~\citep{chen2021summscreen}, which contains two non-overlapping set of TV shows, from different sources. The two sources provide transcripts for two different domains - prime time TV shows and daytime `soap operas', respectively forming the sub-task 3 and 4.   

We share the instructions to access the data for each of the four sub-tasks and associated scripts on our github repo~\footnote{\url{github.com/fladhak/creative-summ-data}}. 

We detail the process of evaluation of these sub-tasks, presenting the metrics as well as the results from the submissions by the shared-task participants alongside several baseline, long-document summarization models. We discuss ways to expand the sub-tasks to include more genres of creative writing for automatic summarization systems in the future.  

\section{Datasets}\label{sec:datasets}

In this section we describe the sources and pre-processing steps taken to curate the data from each sub-task of \DatasetName. The data samples for each of the sub-tasks are available on the shared task github repo. 

\subsection{Sub-Task 1: Summarizing book chapters}\label{ssec:subtask1}
The dataset for sub-task 1 (\textit{BookSumm-chapters}) pairs chapters of novels released as part of Project Gutenberg with corresponding summaries from different online study guides. Source texts for these chapters have been made available in accordance with Project Gutenberg's guidelines.~\footnote{\url{https://www.gutenberg.org/policy/robot_access.html}} For each book-chapter, we provide a link to the online study guide on Web Archive\footnote{\url{https://web.archive.org/}} where the corresponding ground truth summary can be found, for training and validation.

We combine the book titles and the study guide sources used by~\citet{kryscinski2021booksum} and~\citet{Ladhak:20}, and ensure that we remove redundant titles and filter out unreliable study guides. We also filter out book-chapters that were identified as plays (such as those by Shakespeare), as they differ significantly from the other literary genres in the dataset. The book titles in each of our resulting data splits are non-overlapping.

\subsection{Sub-Task 2: Summarizing movie transcripts}\label{ssec:subtask2}
\textit{Scriptbase}~\citep{gorinski2015movie} compiles a corpus of 
1276 movie scripts/screenplays spanning 1909–2013, by crawling relevant web-sites. They pair the scripts with corresponding user-written summaries for the task of summarizing movie transcripts. The movie scripts comprise 
23 different genres, and also include rich information about the multimodal aspects of the various scenes. The authors introduce the task of summarizing movie transcripts, as selecting a chain of scenes that accurately represents a film's story. 
In addition to the original Scriptbase data, we provide a \textit{new} test set containing transcripts and summaries that did not appear in the original Scriptbase release. 

\subsection{Sub-Task 3: Summarizing transcripts from primetime tv-shows}\label{ssec:subtask3}

We utilize the SummScreen dataset~\citep{chen2021summscreen} for the task of summarizing transcripts of episodes from primetime tv-shows. The dataset (\textit{SummScreenFD}) contains community-contributed transcripts for 4348 episodes from 88 shows, gathered from ForeverDreaming (FD).\footnote{\url{transcripts.foreverdreaming.org}} The transcripts are paired with human-written summaries. These transcripts are characterized by long storylines, often involving parallel sub-plots and with great emphasis on character development. For this subtask, we use the data and pre-processing associated with the SCROLLS benchmark~\citep{shaham2022scrolls}, but we provide a \textit{new} test set containing transcripts and summaries not released in either the original SummScreen or SCROLLS datasets. 

\subsection{Sub-Task 4: Summarizing transcripts from daytime `soap-operas'}\label{ssec:subtask4}

The dataset for sub-task 4 is made available in the exact same format as~\ref{ssec:subtask3}, except the transcripts are specific to daytime `soap opera' type shows. Introduced by \citet{chen2021summscreen}, \textit{SummScreenTMS} contains transcripts from 22.5k episodes, from TV MegaSite, Inc. (TMS).\footnote{\url{http://tvmegasite.net/}} Here again we provide a \textit{new} test set containing transcripts and summaries not released in the original SummScreen dataset. 

\subsection{Data Splits}\label{ssec:data-splits}

As noted above, while we used the original data splits for Sub-Task 1, we provided new unseen test inputs for Sub-Tasks 2, 3 \& 4. For \textit{SummScreenFD} \& \textit{SummScreenTMS} we suggested participants use the original test set for validation, and train on the combination of the original train and validation sets. We provide the data splits and some statistics for each sub-task in Table~\ref{tab:data-splits-creativesumm}.

\begin{table*}[!thbp]
    \small
    \centering
    \begin{tabular}{lrrrrrrrrr} 
    \toprule
        \textbf{Dataset} & \textbf{Train} & \textbf{Val}  & \textbf{Test}& \textbf{Coverage} & \textbf{Density} &  \textbf{Comp. Ratio} & \textbf{1-grams} & \textbf{2-grams} \\
    \midrule
    \textit{BookSumm-chapters} & 6754 & 983 & 851 & 0.7062 & 1.4078 & 16.4287 & 0.4456 & 0.8163\\
    \textit{Scriptbase} & 1149 & 127  & 216 & 0.7637 & 1.2624 & 63.2474 & 0.3123 & 0.667 \\
    \textit{SummScreenFD} & 4011 & 337 & 459 & 0.7203 & 1.1138 & 87.5306 & 0.2640 & 0.7066 \\
    \textit{SummScreenTMS} & 20710 & 1793 & 679 & 0.7665 & 1.1954 & 21.0675 & 0.2499 & 0.6969 \\
    \bottomrule
    \end{tabular}
    
    \caption{
    Statistics of the documents in the dataset curated for each sub-task. We report the number of documents in each split, along with the coverage, density and compression ratio b/w the documents and the summaries in the full dataset. We also present the percentage of novel uni and bi-grams present in the reference summaries for documents in each sub-task.
    }
    \label{tab:data-splits-creativesumm}
\end{table*}

\section{Evaluation}\label{sec:evaluation}

We received 18 submissions in total for our shared task, with 
two, eleven, three and two 
submissions for shared tasks 1, 2, 3 and 4 respectively. We describe the metrics used for evaluating the submitted outputs. Participants had 
~12 weeks to sign up for the shared task and train their models, before we released the test set(s) for the different sub-tasks. We allowed another week after releasing the test set for the submission of system outputs. 
We encouraged multiple submissions for each sub-task as long as they reflected different models or approaches.

\subsection{Metrics}\label{ssec:evaluation-metrics}

We used the same metrics for the evaluation of the 
four sub-tasks for easier comparison:

\par \noindent
\textbf{ROUGE} \cite{Lin:04}:
We apply the standard F1 variations of ROUGE-1, ROUGE-2, and ROUGE-L using the original PERL-based implementation.
\par \noindent
\textbf{BERTScore} \cite{Zhang:20}: We compute the precision, recall, and F1 variations of reference-based BERTScore metric. We provide the hash of our BERTScore runs.~\footnote{microsoft/deberta-xlarge-mnli\_L40\_no-idf\_version=0.3.9(hug\_trans=4.20.1)}
\par \noindent
\textbf{LitePyramid (LP)} \cite{zhang-bansal-2021-finding}: This metric has reported state-of-the-art correlations for relevance estimation on news summarization.
This metric is fully automatic and first extracts semantic textual units from the reference summaries and then calculates the entailment score between the model summary and these units.
We report the three-class entailment probability using an entailment model fine-tuned on TAC 2008 \cite{DBLP:conf/tac/2008}. 
%
\par \noindent
\textbf{SummaC} \cite{laban-etal-2022-summac}: We chose this metric as it relies on aggregating sentence-level computations, allowing us to score long input texts. 
We use zero-shot variation of the model, whose NLI component is trained on the VitaminC dataset \cite{schuster-etal-2021-get}.
\par \noindent
\textbf{Summary Statistics}: We report the average length and percentage of novel uni and bi-grams present in the model summaries. We also calculate the extractive coverage and density from \citet{grusky-etal-2018-newsroom}, which measure the extent of the overlap between the summary and input texts.
%
%
\subsection{Baselines}\label{ssec:evaluation-baselines}

For each sub-task, we train three Longformer Encoder-Decoder (LED) \cite{beltagy2020longformer} models that have a maximum input size of 1024, 4096, and 16384, respectively.

\subsection{Results}\label{ssec:evaluation-results}
%
The results for each of the sub-tasks 
are presented in Appendix~\ref{app:shared-task-results}.

The submissions on all tasks surpass the performance of the baseline LED models by a large margin in terms of ROUGE and BERTScore. However, baselines and submissions score poorly in LP and SummaC, with several system scores near zero. These two metrics have been primarily studied within the news domain and with relatively short input, and additional analysis is required to validate these metrics within creative writing and longer input/output summarization. 

Within the \textit{BookSum-chapters} dataset, we find a sizable variation in the length and level of abstraction among systems. We also note that among the tasks, LP and SummaC give the highest scores to the systems here. Notably, we do not see a clear correspondence between the level of abstraction and the SummaC factual consistency score, which has often been observed in the news domain \cite{ladhak-etal-2022-faithful}.

For the \textit{Scriptbase} task, we received many variations of a base model from one team with different performance, although the gap between system statistics is not very large. The ROUGE performance on this dataset is highest among all tasks, although we again note the low performance of all systems according to LP and SummaC.

For the \textit{SummScreenFD} task, the system that performed best consisted of much shorter, fairly abstractive summaries. For the \textit{SummScreenTMS} task, the LED baselines perform very poorly, and we found the output to be largely repetitive, likely due to optimization problems during training. The resulting baselines were much longer and more extractive than the submitted systems. 
The difference in system performance on the two SummScreen tasks (0.29 vs. 0.39 ROUGE-1 on SummScreen FD and TMS, respectively) demonstrates important data differences even within the same domain. 


\section{Related Work}\label{sec:related-tasks}

Although summarization of newswire text has dominated research for over a decade, there have been key efforts at encouraging summarizing research in other domains. \citet{ws-2011-automatic} introduced a workshop on summarizing different genres, media and languages, with the aim of defining new tasks and corporas in these domains. 

\citet{toubia2017summarization} and \citet{toubia2021poisson} highlight the need for summarization of creative documents, by arguing that summaries may serve as a “lubricant” in the market for creative content, making it easier for consumers to decide which information to consume. 


TVRecap~\citep{chen2021tvrecap} introduced a story generation task that requires generating detailed recaps from a brief summary and a set of documents describing the characters involved in the episode. The dataset contains 26k episode recaps of TV shows gathered from fan-contributed websites.


Past work has demonstrated the importance of character analysis for understanding and summarizing the content of movie scripts~\cite{sang2010character, tran2017exploiting}.

With increased focus on characters in creative documents like movies and fictional stories, the task of generating character descriptions using automatic summarization been proposed recently~\cite{zhang2019generating}. The accompanying dataset contains 1,036,965 fictional stories and 942,218 summaries.

While there has been prior work on summarizing short stories~\cite{kazantseva2006approach}, more recent methods in summarizing books utilize the hierarchical structure of documents for understanding the complex inter-dependencies in the text. ~\citet{wu2021recursively} incorporate human feedback along with recursive task decomposition, using summaries of small sections of the book to produce an overall summary at inference time. ~\citet{pang2022long} propose a novel top-down and bottom-up inference framework, which is effective on a variety of long document summarization benchmarks, including books.
\section{Discussion}\label{sec:discussion}


 This workshop proposes the task of summarizing documents containing creative textual content. We consider sub-tasks focusing on the summarization of book chapters, movie scripts and transcripts from TV shows. Each sub-task highlights key challenges in automatic summarization of creative texts in a different genre. We also discuss how other efforts in the literature have attempted to approach this area of research. In its first version as part of COLING 2022, our sub-tasks attracted 18 submissions in total. We present the results from these submissions, along with some baselines and compare the performance of the different systems. 

Summarization of creative texts opens the door for the development of computer-based tools to aid authors and marketers in creative industries~\cite{toubia2021poisson}. Automatic summaries can serve as an important component of screenwriting and book-writing tools, helping grade the quality and gauge interest amongst the consumers~\cite{gorinski2015movie}.

Advances in Creative Summarization will also assist and augment research in other related tasks and areas. Cues from textual summaries and automated content analysis in movie scripts can be helpful in creating movie-image summaries~\cite{tsoneva2007automated}.

 There are two main directions to improve efforts in this direction in the future - incorporating more datasets/sub-tasks that relate to creative summarization, and improving metrics/strategies to effectively evaluate these systems. Other sources on the web, such as the one used by~\citet{zhang2019generating} can be used for harnessing datasets for summarizing creative texts. Similarly, our evaluation scheme can be expanded to include more entity-centric metrics~\cite{chen2021summscreen}, which can be crucial for identifying the presence of key characters in summaries of creative texts.

\bibliography{acl2020}

\appendix
\section{Shared Task Submissions}\label{app:shared-task-results}

The evaluation results for all the submissions to the sub-tasks are presented below in Tables~\ref{tab:sub-task1-results}-\ref{tab:sub-task4-results}. We use the original name of the system outputs as submitted by the participants to identify each submission.
\begin{table*}[ht]
    \centering
    \small
    
    \resizebox{\linewidth}{!}{%
    \begin{tabular}{lrrrrrrrrrrrrr} 
    \toprule

    \textbf{Models} & $\textrm{R}_{1f}$ & $\textrm{R}_{2f}$ & $\textrm{R}_{Lf}$ & $\textrm{BS}_{P}$ & $\textrm{BS}_{R}$ & $\textrm{BS}_{F1}$ & $\textrm{LP}_{p3c}$ & $\textrm{SummaC}_{ZS}$ & $\textrm{Length}$ & $\textrm{Density}$ & $\textrm{Coverage}$ & $\textrm{1-grams}$ & $\textrm{2-grams}$ \\

    \midrule
    & & \multicolumn{11}{c}{\textbf{Baselines}} \\
    \midrule
    $LED_{1024}$ & 0.1413 & 0.0214 & 0.1329 & 0.4318 & 0.4697 & 0.4446 & 0.2684 & 0.2599 & 937 & 1.5149 & 0.7230 & 0.2886 & 0.7094 \\
    $LED_{4096}$ & 0.1537 & 0.0221 & 0.1426 & 0.4232 & 0.4727 & 0.4420 & 0.2642 & 0.4718 & 866 & 2.9900 & 0.7617 & 0.2409 & 0.6096 \\
    $LED_{16384}$ & 0.1487 & 0.0218 & 0.1385 & 0.4282 & 0.4928 & 0.4544 & 0.2221 & 0.4132 & 755 & 2.2112 & 0.7372 & 0.2532 & 0.6541 \\ 
    \midrule
    
    & & \multicolumn{11}{c}{\textbf{Sub-Task 1 submissions}} \\
    \midrule
    MHPK  & 0.2975 & 0.0789 & 0.2833 & 0.5303 & 0.5553 & 0.5410 & 0.1071 & 0.1562 & 1465 & 2.1802 & 0.7735 & 0.3055 & 0.6775   \\
    $LRL\_{NC}$ & 0.2643 & 0.0471 & 0.2436 & 0.4717 & 0.5264 & 0.4955 & 0.0669 & 0.3499 & 3345 & 12.6336 & 0.8336 & 0.1600 & 0.3457 \\

    \bottomrule
    \end{tabular}
    }
    \caption{
    Sub-task 1 results for the baselines and the submissions on the \textit{BookSumm-chapters} dataset.
    }
    \label{tab:sub-task1-results}
    
     \bigskip
    
    \resizebox{\linewidth}{!}{%
    \begin{tabular}{lrrrrrrrrrrrrrrrr} 
    \toprule

    \textbf{Models} & $\textrm{R}_{1f}$ & $\textrm{R}_{2f}$ & $\textrm{R}_{Lf}$ & $\textrm{BS}_{P}$ & $\textrm{BS}_{R}$ & $\textrm{BS}_{F1}$ & $\textrm{LP}_{p3c}$ & $\textrm{SummaC}_{ZS}$ & $\textrm{Length}$ & $\textrm{Density}$ & $\textrm{Coverage}$ & $\textrm{1-grams}$ & $\textrm{2-grams}$ \\

    \midrule
    & & \multicolumn{11}{c}{\textbf{Baselines}} \\
    \midrule
$LED_{1024}$                     & 0.1492 & 0.0146 & 0.1373 & 0.4298 & 0.4238 & 0.4258 & 0.2517 & 0.0000 & 903 & 1.1809 & 0.7021 & 0.3357 & 0.7485 \\
$LED_{4096}$                    & 0.1416 & 0.0130 & 0.1299 & 0.4245 & 0.4137 & 0.4179 & 0.2674 & 0.0000 & 877 & 1.3432 & 0.7311 & 0.3092 & 0.7273 \\
$LED_{16384}$                    & 0.1368 & 0.0125 & 0.1277 & 0.4322 & 0.3924 & 0.4099 & 0.2312 & 0.0000 & 551 & 1.4103 & 0.7266 & 0.3024 & 0.7211 \\
    \midrule
    
    & & \multicolumn{11}{c}{\textbf{Sub-Task 2 submissions}} \\
    \midrule
MovING\_scriptbase            & 0.4144          & 0.0823          & 0.3963          & 0.5163          & 0.5233          & 0.5194           & 0.0356                       & 0.0476              & 729               & 3.4398               & 0.8759               & 0.1621               & 0.4827                \\
UdS\_scriptbase               & 0.4285          & 0.1088          & 0.4125          & 0.5543          & 0.5410          & 0.5474           & 0.0587                       & 0.0323              & 883               & 1.3489               & 0.7778               & 0.2911               & 0.7346                \\
UdS\_base\_bs4\_0.2noisy      & 0.4634          & 0.1158          & 0.4405          & 0.5723          & 0.5680          & 0.5700           & 0.0372                       & 0.0250              & 791               & 1.2722               & 0.7502               & 0.3520               & 0.7636                \\
UdS\_base\_bs4\_Bitfit        & 0.4344          & 0.1091          & 0.4178          & 0.5533          & 0.5402          & 0.5465           & 0.0566                       & 0.0321              & 815               & 1.3475               & 0.7753               & 0.2879               & 0.7301                \\
UdS\_base\_bs4\_Bitfit\_Mix01 & 0.4580          & 0.1162          & 0.4376          & 0.5664          & 0.5601          & 0.5631           & 0.0459                       & 0.0316              & 770               & 1.3226               & 0.7590               & 0.3235               & 0.7457                \\
UdS\_base\_bs4\_Bitfit\_Mix10 & 0.4576          & 0.1158          & 0.4380          & 0.5690          & 0.5620          & 0.5653           & 0.0418                       & 0.0301              & 781               & 1.3194               & 0.7628               & 0.3167               & 0.7459                \\
UdS\_base\_bs4                & 0.4639          & 0.1152          & 0.4408          & 0.5703          & 0.5672          & 0.5686           & 0.0408                       & 0.0269              & 780               & 1.2733               & 0.7513               & 0.3456               & 0.7604                \\
UdS\_base\_bs5\_Bitfit        & 0.4316          & 0.1088          & 0.4151          & 0.5542          & 0.5414          & 0.5475           & 0.0567                       & 0.0309              & 838               & 1.3451               & 0.7774               & 0.2890               & 0.7327                \\
UdS\_base\_bs5\_Bitfit\_Mix01 & 0.4550          & 0.1149          & 0.4344          & 0.5664          & 0.5606          & 0.5634           & 0.0508                       & 0.0322              & 776               & 1.3344               & 0.7621               & 0.3205               & 0.7438                \\
UdS\_base\_bs5\_Bitfit\_Mix10 & 0.4553          & 0.1146          & 0.4353          & 0.5677          & 0.5619          & 0.5646           & 0.0435                       & 0.0295              & 777               & 1.3235               & 0.7596               & 0.3218               & 0.7462                \\
UdS\_base\_bs5                & 0.4604          & 0.1140          & 0.4369          & 0.5699          & 0.5668          & 0.5682           & 0.0410                       & 0.0281              & 783                & 1.2709               & 0.7493               & 0.3485               & 0.7649                \\

    \bottomrule
    \end{tabular}
    }%
    \caption{
    Sub-task 2 results for the baselines and the submissions on the \textit{Scriptbase} dataset.
    }
    \label{tab:sub-task2-results}

    \bigskip
    \resizebox{\linewidth}{!}{%
    \begin{tabular}{lrrrrrrrrrrrrrrrr} 
    \toprule

    \textbf{Models} & $\textrm{R}_{1f}$ & $\textrm{R}_{2f}$ & $\textrm{R}_{Lf}$ & $\textrm{BS}_{P}$ & $\textrm{BS}_{R}$ & $\textrm{BS}_{F1}$ & $\textrm{LP}_{p3c}$ & $\textrm{SummaC}_{ZS}$ & $\textrm{Length}$ & $\textrm{Density}$ & $\textrm{Coverage}$ & $\textrm{1-grams}$ & $\textrm{2-grams}$ \\

    \midrule
    & & \multicolumn{11}{c}{\textbf{Baselines}} \\
    \midrule
$LED_{1024}$                      & 0.1428          & 0.0154          & 0.1236          & 0.4100          & 0.4107          & 0.4052           & 0.0987                       & 0.0559              & 330               & 1.1440               & 0.7148               & 0.3060               & 0.7801                \\
$LED_{4096}$                      & 0.1694          & 0.0209          & 0.1501          & 0.4591          & 0.4752          & 0.4600           & 0.0304                       & 0.1052              & 188               & 1.4378               & 0.7343               & 0.2803               & 0.7314                \\
$LED_{16384}$                     & 0.1514          & 0.0170          & 0.1334          & 0.4485          & 0.4632          & 0.4489           & 0.0304                       & 0.1644              & 192               & 1.5474               & 0.7108               & 0.2904               & 0.7285                \\
    \midrule
    
    & & \multicolumn{11}{c}{\textbf{Sub-Task 3 submissions}} \\
    \midrule

inotum    & 0.2860          & 0.0624          & 0.2529          & 0.5934          & 0.5609          & 0.5750           & 0.0559                       & 0.0272              & 86                & 1.0321               & 0.6664               & 0.3715               & 0.8251                \\
team\_ufal            & 0.2469          & 0.0408          & 0.2300          & 0.5038          & 0.5590          & 0.5285           & 0.0406                       & 0.1282              & 289               & 2.0821               & 0.7127               & 0.2484               & 0.6498                \\
AMRTVSumm & 0.2307          & 0.0303          & 0.2106          & 0.4906          & 0.5344          & 0.5108           & 0.0138                       & 0.024               & 256               & 0.8789               & 0.6137               & 0.4924               & 0.8569                \\

    \bottomrule
    \end{tabular}
    }%
    \caption{
    Sub-task 3 results for the baselines and the submissions on the \textit{SummScreenFD} dataset.
    }
    \label{tab:sub-task3-results}

        \bigskip
        \resizebox{\linewidth}{!}{%
    \begin{tabular}{lrrrrrrrrrrrrrrrr} 
    \toprule

    \textbf{Models} & $\textrm{R}_{1f}$ & $\textrm{R}_{2f}$ & $\textrm{R}_{Lf}$ & $\textrm{BS}_{P}$ & $\textrm{BS}_{R}$ & $\textrm{BS}_{F1}$ & $\textrm{LP}_{p3c}$ & $\textrm{SummaC}_{ZS}$ & $\textrm{Length}$ & $\textrm{Density}$ & $\textrm{Coverage}$ & $\textrm{1-grams}$ & $\textrm{2-grams}$ \\

    \midrule
    & & \multicolumn{11}{c}{\textbf{Baselines}} \\
    \midrule
$LED_{1024}$                            & 0.0727          & 0.0063          & 0.0652          & 0.4100          & 0.4107          & 0.4052           & 0.0304                       & 0.0905              & 984               & 0.9530               & 0.6302               & 0.3781               & 0.8379                \\
$LED_{4096}$                           & 0.0822          & 0.0062          & 0.0753          & 0.4122          & 0.4086          & 0.4057           & 0.0304                       & 0.0837              & 879               & 0.8457               & 0.5822               & 0.4091               & 0.8674                \\
$LED_{16384}$                           & 0.0722          & 0.0047          & 0.0656          & 0.4096          & 0.3916          & 0.3960           & 0.0304                       & 0.0800              & 885               & 0.7544               & 0.5430               & 0.4387               & 0.8931                \\ 

    \midrule
    
    & & \multicolumn{11}{c}{\textbf{Sub-Task 4 submissions}} \\
    \midrule

AdityaUpadhyay & 0.3921          & 0.0909          & 0.3794          & 0.5507          & 0.5550          & 0.5516           & 0.0625                       & 0.1133              & 316               & 1.9436               & 0.7618               & 0.2026               & 0.6688                \\
AMRTVSumm      & 0.3426          & 0.0717          & 0.328           & 0.5385          & 0.5318          & 0.5347           & 0.0152                       & 0.0499              & 259               & 1.1024               & 0.7291               & 0.3514               & 0.7931                \\

    \bottomrule
    \end{tabular}
    }%
    \caption{
    Sub-task 4 results for the baselines and the submissions on the \textit{SummScreenTMS} dataset.
    }
    \label{tab:sub-task4-results}
\end{table*}

%
%
%


\FloatBarrier
\end{document}